\documentclass[11pt]{article}
\usepackage{coling2018}
\usepackage{times}
\usepackage{url}
\usepackage{latexsym}
\usepackage[utf8]{inputenc}

\usepackage{url}
\usepackage{booktabs,arydshln}
\usepackage{multirow,bigdelim}

\usepackage{xspace}

\usepackage{tikz}
\usetikzlibrary{shapes.geometric}
\usepackage{pgfplots}
\pgfplotsset{compat=1.5}

\let\citet\newcite

\makeatletter
\def\adl@drawiv#1#2#3{%
        \hskip.5\tabcolsep
        \xleaders#3{#2.5\@tempdimb #1{1}#2.5\@tempdimb}%
                #2\z@ plus1fil minus1fil\relax
        \hskip.5\tabcolsep}
\newcommand{\cdashlinelr}[1]{%
  \noalign{\vskip\aboverulesep
           \global\let\@dashdrawstore\adl@draw
           \global\let\adl@draw\adl@drawiv}
  \cdashline{#1}
  \noalign{\global\let\adl@draw\@dashdrawstore
           \vskip\belowrulesep}}
\makeatother

\newcommand\uoi{\texttt{UoI}}
\newcommand\clp{\texttt{clp}}
\newcommand\clpa{\texttt{clp-all}}
\newcommand\neural{\texttt{neural}}
\newcommand\stanford{\texttt{stanford}}
\newcommand\crf{\texttt{crf}}
\newcommand\gnb{\texttt{G\&B}}
\newcommand\mono{\texttt{mono}}
\newcommand\crfm{\texttt{crf-mod}}

\newcommand{\ndaff}{\scalebox{1.0}{\ensuremath{^\spadesuit}}\xspace}

\newcommand{\ioaff}{\scalebox{1.0}{\ensuremath{^\diamondsuit}}\xspace}

\newcommand{\bcaff}{\scalebox{1.0}{\ensuremath{^\clubsuit}}\xspace}

\title{Part-of-Speech Tagging on an Endangered Language: \protect\\ a Parallel Griko-Italian Resource}

\author{
Antonis Anastasopoulos\ndaff ~ Marika Lekakou\ioaff ~ Josep Quer\bcaff ~ Eleni Zimianiti\ioaff\\
\textbf{Justin DeBenedetto}\ndaff ~ \textbf{David Chiang}\ndaff\\
 \ndaff{}Department of Computer Science and Engineering, University of Notre Dame\\
 \ioaff{}Department of Philology, University of Ioannina\\
 \bcaff{}ICREA \& Department of Translation and Language Sciences, Universitat Pompeu Fabra\\
 {\tt aanastas@nd.edu, mlekakou@cc.uoi.gr, josep.quer@upf.edu}
}

\begin{document}
\maketitle
\begin{abstract}
Most work on part-of-speech (POS) tagging is focused on high resource languages, or examines low-resource and active learning settings through simulated studies. We evaluate POS tagging techniques on an actual endangered language, Griko. 
We present a resource that contains 114 narratives in Griko, along with sentence-level translations in Italian, and provides gold annotations for the test set.
Based on a previously collected small corpus, we investigate several traditional methods, as well as methods that take advantage of monolingual data or project cross-lingual POS tags. We show that the combination of a semi-supervised method with cross-lingual transfer is more appropriate for this extremely challenging setting, with the best tagger achieving an accuracy of~$72.9\%$. 
With an applied active learning scheme, which we use to collect sentence-level annotations over the test set, we achieve improvements of more than~$21$ percentage points. 
\end{abstract}

\section{Introduction}

\blfootnote{
    \hspace{-0.65cm}  
    This work is licensed under a Creative Commons 
    Attribution 4.0 International License.
    License details:
    \protect{\url{http://creativecommons.org/licenses/by/4.0/}}
}

Most natural language processing (NLP) applications have been developed for and tested on only a handful of languages. The majority of the world's languages are under-represented in the field, mostly due to the lack of proper resources. 
Endangered languages pose additional problems, as the lack of resources is further exacerbated by the lack of standard orthography. 
Therefore, there is an obvious need for both resources and technologies adapted to languages of under-represented communities. 
Especially in the case of endangered languages, computational approaches can be used to scale and accelerate documentation and revitalization efforts.

For the purposes of this study, we focus on Griko, a Greek dialect spoken in southern Italy, in the Grecìa Salentina area southeast of Lecce.\footnote{A discussion on the possible origins of Griko can be found in the paper by \citet{manolessou2005greek}.} 
There is another endangered Italo-Greek variety in southern Italy spoken in the region of Calabria, known as Grecanico or Greco. Both languages, jointly referred to as \textit{Italiot Greek}, were included as seriously endangered in the UNESCO \textit{Red Book of Endangered Languages} in~1999.
Griko is only partially intelligible with modern Greek, and unlike other Greek dialects, it uses the Latin alphabet. Less than~20,000 people (mostly people over 60 years old) are believed to be native speakers \cite{horrocks2009greek,douri2015griko}, a number which is quite likely an overestimation \cite{chatzikyriakidis2010clitics}. 

In general, the lack of annotated resources can be addressed through several directions. The obvious first one is the collection of human annotations, an expensive and time-consuming process. Another option is to collect additional monolingual data and use them in weakly-supervised approaches. Finally, one could use methods that transfer knowledge across languages, in which case they could focus on collecting translations or bilingual data. With this paper we attempt to quantify how effective each of these directions can be at the beginning of building applications for an endangered language.

We present a corpus of Griko narratives, first collected in the beginning of the 20th century, along with their Italian translations, suitable for computational research on the language.\footnote{Our corpus is available for download here: \url{https://bitbucket.org/antonis/grikoresource}} 
We focus on the task of part-of-speech (POS) tagging, an important subcomponent of many downstream NLP applications, although our long-term goal is to annotate the whole corpus with morphosyntactic tags. 
Three linguists were tasked with creating gold-standard annotations for a portion of the resource, which we use as a test set in our experiments.
Subsequently, our corpus includes human annotations as well as monolingual data and translations, allowing us to explore all approaches.

Based on another small Griko corpus with POS annotations, we explore computational methods for annotating our collected resource with POS tags.
We evaluate several commonly used models for POS-tagging. 
We start with a traditional feature-based cross-entropy tagger, a Conditional Random Field (CRF) tagger, and a neural bi-LSTM tagger. 
We further experiment with methods that take advantage of additional monolingual data, and with methods that project cross-lingual tags through alignments. 
In addition, we are the first to combine the two latter approaches for POS-tagging in a low-resource setting, and we show that their combination achieves higher accuracy.
Finally, we study how additional human annotations can be incorporated through an applied active learning scenario. In line with previous work, we show that this greatly improves the tagging accuracy.

Our contributions are three-fold: first, we aim to provide coarse insights into what type of annotations and resources are more effective at the early stages of developing a tagging tool for an endangered language.
Second, we hope that the release of our resource will spark further interest in the computational community, so that previous, and future, methods are tested in actual endangered language settings.
Finally, we benchmark the accuracy of several models on our dataset, and test them in a traditional setting and in a \textit{transduction} setting where we also use the translations of the test set. We also evaluate all methods in an active learning scenario, showing that different approaches are suitable for different amounts of annotated data.

When only 360 annotated sentences are available, the best method is the one that combines them with both cross-lingual projections and monolingual data, achieving an accuracy of~72.9\%. 
As the active learning scheme unfolds, however, we show that there is no need for either semi-supervised nor transfer learning approaches: a simpler feature-based CRF model achieves the highest scores, with more than~94\% accuracy.

\section{Resource}
\begin{table*}[t]
    \centering
    \begin{tabular}{c|cccccc}
    \toprule
         & \multirow{2}{*}{Stories} & \multirow{2}{*}{Sentences} & \multicolumn{2}{c}{Griko} & \multicolumn{2}{c}{Italian} \\
         & & & Types & Tokens & Types & Tokens \\
    \midrule
        train & 104 & 9.2k & 13.5k &197.6k& 10.6k &169.7k \\
        test & 10 & 885 & 2.4k & 14.0k & 2.3k & 13.1k\\
    \midrule
        all & 114 & 10.1k & 14.1k & 211.6k & 11.0k & 182.7k\\
    \bottomrule
    \end{tabular}
    \caption{Statistics on our collected Griko-Italian resource.}
    \label{tab:resource_stats}
\end{table*}

Resources in Griko are very scarce. The German scholar Gerhard
Rohlfs pioneered research on Griko and composed the first grammar of the language \cite{rohlfs1977grammatica}, also heavily influencing the subsequent grammar created by \citet{karanastasis1997grammatiki}. Although the language has been further studied, almost no corpora are available for linguistics research.

The only Griko corpus available online\footnote{\url{http://griko.project.uoi.gr}} \cite{grikodatabase} consists of about~20 minutes of speech in Griko, along with text translations into Italian. The corpus (henceforth \uoi{} corpus, as it is hosted at the University of Ioannina, Greece) consists of~330 mostly elicited utterances by nine native speakers, annotated with transcriptions, morphosyntactic tags, and glossing in Italian.

The most noted Griko scholar is Vito Domenico Palumbo (1854--1928) who made the first serious attempts to create a literary Griko for the dialect of Calimera (the most populous of the nine remaining communities where Griko is still spoken), based on modified Italian orthography. Salvatore Tommasi and Salvatore Sicuro then edited and published Palumbo's manuscripts \cite{palumbo1998io,palumbo1999itela}, a part of which we now make available for computational and linguistic research.

After scraping from their website\footnote{\url{https://www.ciuricepedi.it}}~114 narratives that Palumbo had collected, along with their Italian translations, we removed all HTML markup and normalized the orthography: 
we substituted all curly quotes and apostrophes with simple ones, and substituted the vowels with circumflex (â, ô, û) that were used in a few contractions with the more common accented vowel--apostrophe combination (à', ò' ù'). Using the Moses tools \cite{koehn2007moses} with the Italian settings, we lowercased and tokenized our parallel dataset. For completeness purposes, we also make available the untokenized and proper-case versions of the corpus. The statistics of the resource are shown in Table~\ref{tab:resource_stats}. 

We chose the first $10$ narratives to be our test set, as they correspond to about~$10\%$ of all sentences. 
The rest of the narratives are treated as a monolingual or parallel resource to be leveraged.
The test set was, in addition, hand-annotated by linguists: they corrected any tokenization errors that were introduced by the automatic process (for example, regarding the use of the apostrophe) and produced POS tags for every test sentence.

For every narrative, one of the linguists was presented with the produced output of the tagger, and proceeded to correct it. In order to ensure the quality of the annotations, a second linguist was then presented with the result of the work of the first linguist and tasked with correcting it, until all disagreements were resolved. Although it significantly slowed down the annotation process, we hope that this scheme ensured the quality of our annotations.

\subsection{Differences from previous Griko resources} 
\begin{table*}[t]
    \centering
    \begin{tabular}{lclcll}
    \toprule
    tag & frequency & tag & frequency & tag & frequency \\
    \midrule
    V \textit{(verb)} & 24.4 & Prt \textit{(particle)} & 2.2  & Adv+Adv & 0.4 \\
    PUNCT \textit{(punctuation)} & 18.3 & P+D & 1.8  & X \textit{(other)} & 0.3 \\
    Pr \textit{(pronoun)} & 12.5 & P \textit{(adposition)} & 1.6 & V+Pr & 0.3\\
    N \textit{(noun)} & 11.6 & Adj \textit{(adjective)} & 1.2 & P+P & 0.1 \\
    C \textit{(complementizer)} & 11.4  & Num \textit{(numeral)} & 0.7 & Pr+Pr & 0.1 \\
    D \textit{(determiner)} & 7.2 & N+Pr & 0.6 & C+Pr & 0.1  \\
    Adv \textit{(adverb)} & 5.0 & V+C & 0.4 & & \\
    \midrule
    \multicolumn{5}{l}{Adv+P, Adv+Pr, Adv+Prt, Adj+Pr, Prt+N, Prt+Pt, D+N, Adv+Adv+Prt, Adv+Adv+Pr} & $<0.1$\\
    \bottomrule
    \end{tabular}
    \caption{List of tags and their frequency in the annotated test part of the corpus.}
    \label{tab:taglist}
\end{table*}

\paragraph{Orthography} Griko has never had a consistent orthography. The transcriptions in the \uoi{} corpus are based on orthographic conventions found in the few textual resources such as the local magazine \textit{Spìtta}, that closely follow conventions adopted in Italian, aiming to be familiar to the speakers of the language. This non-standardization of the orthography leads to variations in the transcription of the same words.

In addition, we find that the word segmentation in our collected narratives follows more the concept of a phonological word. As a result, words that are segmented in the \uoi{} corpus, in our narratives are often fused in a single token. The most common case that also appears in both Italian and Greek, is the contraction of prepositions and subsequent articles, such as the Italian \textit{alla} or the Greek \textit{$\sigma\tau\eta$ (sti)} `to the.Fem'. Other examples of word fusion that is not permitted in either Italian or Greek but appear in our narratives are nouns and possessive pronouns, or adverbs with other adverbs or prepositions. A direct result of this phenomenon is that annotating such tokens with single POS tags does not capture all of the necessary information.

Therefore, we chose to annotate such words with multiple POS-tags, effectively making our tag dictionary the superset of the universal tagset. The final tags that appear in practice in our corpus, and their respective frequencies, are listed in Table~\ref{tab:taglist}. Examples of fused words and their glosses and associated tags are shown in Table~\ref{tab:examples}.

\paragraph{Phonosyntactic Gemination} One important difference is that the \uoi{} corpus explicitly annotates the phenomenon of \textit{raddoppiamento	 fonosintattico} (phonosyntactic gemination, or doubling of the initial consonant of the word in certain contexts) with a hyphen that separates the two words. 
The transcriptions that we collected do not mark for this phenomenon. The two words are often fused into a single token, and the doubling is not always present. 
For example, both following types appear in our corpus: \textit{aderfòmmu} and \textit{aderfòmu} `my brother'.

Furthermore, the \uoi{} corpus also uses apostrophes to	mark word boundaries within which the \emph{raddoppiamento fonosintattico} takes place.
The use of apostrophes in our collected narratives is more loose. They are used both to mark elision/apocope, stress, as well as what it seems to be instances of raddoppiamento fonosintattico. This poses further issues that are discussed in the next paragraph.

\begin{table}[t]
    \centering
    \begin{tabular}{l|c|c|c|c|c|c}
    \toprule
        word: & \textit{stì} & \textit{mànassu} & \textit{cikau} & \textit{ènna} & \textit{vàleti} & \textit{pànuti}\\
        morphemes: & \textit{s[e]-tì} & \textit{màna-su} & \textit{ci-kau} & \textit{è-na} & \textit{vàle-ti} & \textit{pànu-ti}\\
        POS tag: & P+D & N+Pr&Adv+Adv & V+C & V+Pr & Adv+Pr\\
        gloss: & to-the.Fem.SG & mother-your.SG &there-down & have-COMP & put-her & on-her\\
        translation: & `to the' & `your mother' & `down there' & `will' & `put her' & `on her'\\
    \bottomrule
    \end{tabular}
    \caption{Examples of fused types that receive multiple tags in our annotation. The first example is a common preposition-determiner contraction, while the second and last example denote the common fusion of pronouns that follow nouns or adverbs. Notice the doubled consonants in the second and fourth instance, due to  \textit{raddoppiamento fonosintattico}.}
    \label{tab:examples}
\end{table}

\paragraph{Code Switching} There are three languages present in the region of Salento: the regional variety of Italian, the Italo-Romance dialect of Salentino, and Griko.\footnote{See \cite{golovko2013salentino} for a broader overview of the linguistic diversity in the Salento area.} In modern day all members of the Griko community are bilingual or trilingual. The generations before the Second World War are considered to have been predominantly monolingual, and our narratives were collected at that time, around the beginning of the 20th century. However, elements of Salentino do appear in the narratives, either as passing words, or as full sentences, mostly in dialogue turns. Note that resources on Salentino are also extremely scarce if not non-existent. 

In order to deal with such examples, we decided to distinguish two scenarios. Tag switching or intra-sentential switching instances were fully annotated. So, any Salentino words or phrases that appear \textit{within} a Griko sentence, are used for training and evaluation. However, in the few cases where we encounter full sentences in Salentino, we opt to not use them for training or evaluation. Such sentences are marked with distinctive tags in the released corpus. Note that the \uoi{} corpus does not include any non-Griko words or phrases. An extensive study of the code-switching phenomena that occur in our corpus is left for future work.

The following is an example of usage of a Salentino phrase (italicized) within a Griko sentence, taken from story~4. Note that there exists a Griko word for `olive oil', namely \textit{alài} or \textit{alàdi}, as well words for `good', namely \textit{kalòn} or \textit{brao}. However, the Salentino phrase \textit{oju finu} `fine oil' is chosen:

\begin{center}
\begin{tabular}{lllll}
    leo & ti & vastò & \textit{oju} & \textit{finu}\\
    say-1SG & COMP & hold-1SG & oil & fine \\
    V & C & V & N & Adj\\
    \multicolumn{5}{l}{`I say that I have good olive oil'}
\end{tabular}
\end{center}

\paragraph{Tokenization} The \uoi{} corpus has been carefully crafted to make sure that word boundaries are clearly denoted by spaces or hyphens. This unfortunately is not the case in our collected narratives. The ``loose" use of apostrophes complicates the work of the tokenizer. We chose to tokenize all apostrophes as a single token, except for the cases of known elisions that were present in previous corpora, such as the case of the conjunction \textit{c' (ce)} `and'. 
In addition, in the manually annotated test set, the linguists corrected any clear tokenization issues regarding the apostrophe. 

\paragraph{Stress Marking} In the \uoi{} corpus, all words with two or more syllables have a diacritic mark to indicate the location of stress. However, the resources that we collected are not consistent in the use of such a diacritic. Its use is, besides, not standardized and not well studied. Although in most cases such a diacritic is used, there are several instances of polysyllabic words that have no stress marks.

\subsection{Metadata}

We further provide as much information as possible for each narrative, in the form of metadata. This includes the original url of the narrative, the title of the narrative in Griko and its translation in Italian. Whenever they were reported (more than~95\% of the narratives) we include the location where the narrative was collected, and we anticipate that further analysis could possibly reveal any regional variations. The vast majority of the stories were naturally collected in Calimera, the largest village and the center of the Griko community, but the resource also includes 10 stories collected in Martano, as well as stories collected in Corigliano and Martignano, two smaller villages. 
We also include information about the date that a story was collected, as well as the narrator of the story. There are a total of~37 different narrators, while the 10 stories from Martano were retrieved from anonymous manuscripts. There are also~11 stories where the narrator is not known. Two thirds of the stories were narrated by women, while~15\% of the narrators were male. The oldest manuscript dates back to~1883, while the most recent story was collected in~1998.
We hope that this additional information will further allow us to investigate morphosyntactic phenomena in relation to their temporal or location context, but this is left as future work.

\section{Related Work}

POS tagging is a very well studied problem; probabilistic models like Hidden Markov Models and Conditional Random Fields (CRF) were initially proposed \cite{lafferty2001conditional,toutanova2003feature}, with neural network approaches taking over in recent years \cite{mikolov2010recurrent,huang2015bidirectional}. 

The use of parallel data for projecting POS tag information across languages was introduced by \citet{yarowsky2001inducing}, and further improved at a large scale by \citet{das2011unsupervised} who used graph-based label propagation to expand the coverage of labeled tokens. \citet{tackstrom2013token} used high-quality alignments to construct type and token level dictionaries. In the neural realm, \citet{zhang2016ten} used only a few word translations in order to train cross-lingual word embeddings, using them in an unsupervised setting. \citet{meng2017model}, on the other hand, used parallel dictionaries of~20k entries along with~20 annotated sentences.

Most of the previous approaches are rarely tested on under-represented languages, with research on POS tagging for endangered languages being sporadic. \citet{ptaszynski2012part}, for example, presented an HMM-based POS tagger for the extremely endangered Ainu language, based on dictionaries,~12 narratives (yukar), using one annotated story~(200 words) for evaluation. To our knowledge, no other previous work has extensively tested several approaches on an actual endangered language.

The lack of high quality annotated data led to approaches that attempt to use monolingual resources in a semi-supervised setting. Notably, \citet{garrette2013learning} used about~200 annotated sentences along with monolingual corpora improving the accuracy of an HMM-based model.
They tested their model on two low-resource African languages, Kinyarwanda and Malagasy and they found that in this time-constrained scenario type-level annotation leads to slightly higher improvements than token-level annotation, increasing the accuracy of their taggers to slightly less than~80\%. Similar conclusions were reached in \citet{garrette2013real}: 4 hours of annotation are more wisely spent if annotating at the type-level, provided there exist additional raw monolingual data.
This line of work adequately addressed the question of what labeled data are preferable when there is (exceptionally) restricted access to annotators.

However, language documentation neither is nor needs to be restricted to such minimal amounts of annotation work. In addition, recently proposed endangered language documentation frameworks \cite{bird-EtAl:2014:W14-22} advocate the collection of translations \cite{bird-EtAl:2014:Coling} which render the resource interpretable.
In the case of our resource, we argue that the translations are enough for providing type-level supervision. Possibly, this is only feasible because the two languages belong in the same family \emph{and} have been in contact for centuries, so care needs to be taken with the application of this claim.

\section{Part-of-Speech Tagging}
\label{sec:pos}
\begin{table}[t]
    \centering
    \begin{tabular}{cccc}
    \toprule
        \multirow{3}{*}{Model} & \multicolumn{3}{c}{Data} \\
        & \multicolumn{2}{c}{\textit{no transduction}} & \textit{transduction} \\
        & \uoi{} & +\clp & +\clpa\\
    \midrule
        \texttt{stanford} & 62.90 & 67.10 & 67.11 \\
        \texttt{crf} & 57.79 & 59.12 & 59.26\\
        \texttt{crf-mod} & 67.52 & 62.89 & 66.50\\
        \texttt{neural} & 45.27 & 53.24 & 58.50\\
    \midrule
        & \uoi+\mono & +\clp & +\clpa\\
    \midrule
        \gnb & 71.67 & \textbf{72.92} & 72.07\\
    \bottomrule
    \end{tabular}
    \caption{The best performing model is the one that combines semi-supervised learning with cross-lingual projected tags  (\gnb+\clp{}). All models except for \crfm{} benefit from transfer learning through alignments (+\clp{}).
    Transduction does not significantly affect performance, except for the \neural{} model.}
    \label{tab:tagresults}
\end{table}

First, we construct a mapping of the tags of the \uoi{} corpus to the Universal Part-of-Speech tagset \cite{petrov2012universal}. This mapping is available as part of the complementary material of our resource.

Starting with the tagged \uoi{} corpus, we can use several methods to train a tagger, which we use as baselines.
We use the Stanford Log-linear POS-tagger \cite{toutanova2003feature} (henceforth \stanford{}), trained and tested with the default settings. We also test a simple feature-based CRF tagger (henceforth \crf), using the implementation of the \texttt{nltk} toolkit \cite{bird2004nltk}. We extended the implementation to also use prefix and suffix features of up to 4 characters, along with bigram and trigram features.\footnote{Our extensions will be submitted to the \texttt{nltk} codebase.} We will refer to this method as \crfm{}.

Finally, we also investigate the use of a simple neural model. It uses a single bi-LSTM layer to encode the input sentence, and it outputs tags after a fully connected layer applied on the output of the recurrent encoder, as was described in \citet{lample2016neural}. The model is implemented in DyNet\footnote{We will make the code available online.} \cite{neubig2017dynet}, with input embedding and hidden sizes of 128, and output (tag) embedding size of 32. It is trained with the Adam optimizer with an initial learning rate of $0.0002$ and for a maximum of 50 epochs. We select the best model based on the performance on a small dev set of 40 sentences that we sampled randomly from the training set.

The tagging performance of all methods is shown in the first column of Table~\ref{tab:tagresults}. We find that the \crfm{} model is the best baseline model. With such few data to train on, both the \crf{} and the \neural{} model do not perform well. 
The bi- and tri-gram features that the \crfm{} model uses are very sparse, while the \neural{} model has to deal with a very large number of unknown words, as discussed below in the Analysis subsection.

In line with previous work, we find that semi-supervised training achieves better results in such low-resource settings. We exploit all the narratives that we collected by treating them as an additional monolingual corpus, used in the framework proposed by \citet{garrette2013learning}. This approach (henceforth \gnb) significantly improves upon all baselines, achieving an accuracy of~71.67\% in the test set, an improvement of more than~4 percentage points.

\paragraph{Cross-lingual projected tags} So far, our results have not used the Italian translations of our resource. We can follow a procedure similar to the one of \citet{tackstrom2013token}, and extract word alignments from the Griko-Italian parallel data of the training set. We use a pre-trained Italian tagger\footnote{\url{http://elearning.unistrapg.it/TreeTaggerWeb/TreeTagger.html}} in order to tag the Italian side, and we map those tags to the universal tagset. We can then project the tags of the Italian tokens to the aligned Griko ones.\footnote{The type-level projections are also provided with the Supplementary Material.} For the cross-lingual projected tags, we found that in practice type-level predictions work better, and thus we only report results with such models. 
The tags of the Italian side of our resource, the Griko-Italian alignments, and the cross-lingual POS projections on Griko types are available through the complementary material of our resource.

Augmenting the training set with the type-level projected tags (\clp{} in Table~\ref{tab:tagresults}), we achieve improvements for all models, 
except \crfm{}. The \crfm{} method uses sparser features and is more prone to errors due to the noise of the projections.
The best performance is achieved when we combine the projected tags, as type-level supervision, with the \gnb{} method that leverages monolingual data. Their combination achieves the best overall performance, with an accuracy of~72.9\%, a significant improvement over all other methods. As far as we know, this is the first time that cross-lingual projected tags are combined with the method of \citet{garrette2013learning}.

\paragraph{Transduction} An additional approach that needs to be studied is the transductive approach. Since we have translations both for the training and the test sets, we can extract word alignments and project POS tags also for the test set. The results of the transductive approach using cross-lingual projected tags from all the data that we have are shown in the third column of Table~\ref{tab:tagresults} (under \clpa).

We find that most methods benefit from the transductive approach, with the \stanford{} and \crf{} methods exhibiting minimal improvements, while the \neural{} method improves significantly by about~5 percentage points as now there are even less out-of-vocabulary words in the input. The \crfm{} method improves over the \uoi\texttt{+}\clp{} version, but still does not surpass the \uoi{} only version. The only method that does not benefit from the transduction setting is the \gnb{} method, where the performance drops. 

An additional transductive step that can be taken with the \gnb{} method is to also add the test set as part of the monolingual data that it uses. However, including the test set in the monolingual data also resulted in a drop in performance. Using all monolingual data along with the train-only cross-lingual types (\clp) leads to accuracy around ~69.9\% (a drop of~3 points from the best model), while using all monolingual data with \clpa{} leads to a drop of another~1.4 points, to an accuracy of only 68.5\%, which however is still better than all other taggers. These accuracy drops are probably justifiable, since \gnb{} was not developed under a transductive assumption.

\paragraph{Analysis}

It is worth noting that our choice of using combined tags for fused/contracted words means that our training sets, under all settings, do not contain all tags that we encounter in the test set. 
The tagset of the \uoi{} corpus only had 14 tags (the 12 universal ones plus \texttt{P+D} and \texttt{C+Pr}), indicative of its small size. As more narratives were annotated, the size of the necessary tagset increased to the final 29. 
However, the additional tags that we had to use are rather rare and do not severely affect the performance of our models. The tags that are present in the \uoi{} corpus in fact account for~96.7\% of all target tags in the test set, a value that could be considered as a skyline for all methods.

The explanation of our models' performance lies in vocabulary coverage. The \uoi{} corpus only includes~46.6\% of the test set tokens (8.9\% of the test set types). The augmented training set with type-level projections increases those numbers to~48.7\% of test tokens and~14.8\% of test types.
Even though we restrict ourselves to high quality alignments,\footnote{An alignment is used if either its probability is~1, or its probability is higher than~0.9 \textit{and} the frequency of both tokens is higher than 5. Relaxing those conditions leads to worse performance due to noise.} we are able to project tags to~3870 types (3911 in the transductive scenario), an amount higher than the amount of tags that a trained linguist can produce within four hours of annotation \cite{garrette2013real}.

The \gnb{} method deals with the vocabulary coverage issue by introducing a tag dictionary expansion as a first step. They use a label propagation algorithm ---Modified Adsorption \cite{talukdar2009new}--- in order to spread labels between related items. 
In our framework, the cross-lingual projected tags provide labels for a subset of the types, in a way similar that an annotator would, partially alleviating the difficulty of the method's first step.
This leads to less noise in the created tag dictionary, leading to increased accuracy. 
Note that, out of the cross-lingual projected tags that correspond to types that appear in the test set (about~10\% of the test set types, in the \textit{no transduction} setting), more than~65\% were correctly projected.

\section{Active learning}
\label{sec:active}

\begin{table}[t]
    \centering
    \begin{tabular}{clccccc}
    \toprule
        \multirow{2}{*}{Iteration} & \multirow{2}{*}{Narrative} & \multicolumn{2}{c}{Best accuracy} & \multirow{2}{*}{$\Delta$}  & \multicolumn{1}{c}{Accuracy} & \multirow{2}{*}{Best method}\\
        & & without AL & with AL & & on story 9 & \\
    \midrule
        1 & story-1 & 77.89 & --- & 0.0 & 78.13 & \multirow{4}{*}{\gnb+\clp{}}\\
        2 & story-8 & 72.76 & 78.48 & 5.72 & 82.12 \\
        3 & story-7 & 75.07 & 85.17 & 10.10 & 83.57\\
        4 & story-10 & 70.88 & 79.98 & 9.10 & 85.08\\
    \cdashlinelr{4-7}
        5 & story-5 & 72.26 & 82.34 & 10.08 & 88.21 & \multirow{2}{*}{\crfm+\clp{}}\\
        6 & story-4 & 74.03 & 86.30 & 12.27 & 90.32\\
    \cdashlinelr{4-7}
        7 & story-3 & 72.48 & 89.67 & 17.19 & 92.13 & \multirow{3}{*}{\crfm}\\
        8 & story-6 & 74.67 & 91.80 & 17.13 & 93.64\\
        9 & story-2 & 70.78 & 92.67 & 21.89 & 94.17\\
        10 & story-9 & 72.97 & 94.17 & 21.20 & ---\\
    \bottomrule
    \end{tabular}
    \caption{Tagging accuracy for each test narrative with and without active learning. We obtain significant improvements (shown in the $\Delta$ column) by adding each annotated narrative to the training set before retraining and tagging the next narrative. The last two columns further outline the improvements from active learning, showing performance on the last and longest narrative of the test set (story-9) in each iteration, also showing the best method in each iteration. The impact of using monolingual data and type-level cross-lingual projections disappears when more training data are available.}
\label{tab:narratives}
\end{table}

We further explored the use of active learning while tagging our test set. Our active learning scheme is as follows: We first sorted the test set narratives according to length, and starting only with the \uoi{} corpus, we trained all taggers, producing annotations for the first story of the test set. After the corrections on the annotation of each narrative were completed, it was added as gold training data and the taggers were re-trained. For each subsequent story, the linguists were provided with the output of the tagger that achieved the highest accuracy in the previous iteration.

The main reason why  we decided to follow this narrative-level active learning scheme instead of collecting type-level annotations is that a noisy corpus is not very helpful for linguistics research; at least some part of the resource should have to be checked for quality and accuracy by hand. 
In addition, the translations of the narratives can provide such information, as we already showed in the previous section.
As we expand the coverage of our POS annotations over the whole corpus, we will explore other methods for selecting the types or sentences to be annotated through an active learning scheme.

The results, per narrative, with and without active learning, in the order that they were annotated by our linguists (from the shortest narrative to the longest) are outlined in Table~\ref{tab:narratives}. The results for each narrative in the active learning scenario (``with AL" column) report the best performing model that is trained on the concatenation of the \uoi{} corpus and all the stories that were annotated in previous iterations. It is clear that the performance of the taggers improved continuously, as we added more training data. This is further outlined by each iteration's tagging accuracy on story 9, the last and longest narrative of the test set. Of course, when a narrative is added in the training set, it is then excluded from the test set, and the performance is reported on the rest of the narratives.

\begin{figure*}
    \centering

\pgfplotstableread[row sep=\\,col sep=&]{
data & crf & crfj & neural & stanford & gnb-train & gnb-all & crfjtr & gnb \\
370 & 59.26 & 67.52 & 58.50 & 67.11 & 72.92 & 72.07 & 66.50 & 72.92 \\ 
396 & 62.52 & 74.11 & 65.42 & 72.41 & 76.49 & 77.61 & 74.52 & 77.61 \\
423 & 67.30 & 79.33 & 72.23 & 78.02 & 82.30 & 81.52 & 79.94 & 82.30 \\
473 & 71.24 & 82.48 & 75.27 & 80.46 & 82.87 & 83.00 & 81.77 & 83.00 \\
526 & 72.99 & 84.25 & 77.72 & 83.24 & 83.90 & 84.83 & 84.73 & 84.83 \\
609 & 75.03 & 87.09 & 81.50 & 85.62 & 85.26 & 86.06 & 87.52 & 86.06 \\
715 & 77.78 & 89.87 & 85.55 & 88.86 & 85.97 & 87.24 & 90.32 & 87.24 \\
840 & 80.08 & 91.72 & 87.85 & 90.69 & 86.92 & 87.43 & 91.49 & 87.43 \\
975 & 81.39 & 93.31 & 89.40 & 92.42 & 87.22 & 88.07 & 92.29 & 88.07 \\
1105 & 81.96 & 94.17 & 90.55 & 92.86 & 87.93 & 89.01 & 93.07 & 89.01 \\
}\systemdata

\pgfplotsset{width=\textwidth,height=6cm,compat=1.5}
\pgfplotsset{every tick label/.append style={font=\tiny}}

\pgfplotscreateplotcyclelist{my black white}{%
solid, every mark/.append style={solid, fill=gray}, mark=*\\%
dotted, every mark/.append style={solid, fill=gray}, mark=square*\\%
densely dotted, every mark/.append style={solid, fill=gray}, mark=otimes*\\%
loosely dotted, every mark/.append style={solid, fill=gray}, mark=triangle*\\%
dashed, every mark/.append style={solid, fill=gray},mark=diamond*\\%
loosely dashed, every mark/.append style={solid, fill=gray},mark=*\\%
densely dashed, every mark/.append style={solid, fill=gray},mark=square*\\%
dashdotted, every mark/.append style={solid, fill=gray},mark=otimes*\\%
dasdotdotted, every mark/.append style={solid},mark=star\\%
densely dashdotted,every mark/.append style={solid, fill=gray},mark=diamond*\\%
}

\begin{tikzpicture}
    \begin{axis}[
            ymajorgrids=true,
            yminorgrids=true,
            legend style={at={(0.5,1.2)},
                anchor=north,legend columns=5},
            cycle list name=black white,    
            xtick=data,
            ytick={60,70,80,90,100},
            tick pos=left,
            ymin=56,ymax=100,
            xmin=360, xmax=1120,
            ylabel={Accuracy},
            xlabel={Annotated Sentences},
        ]
        \addplot+ table[x=data,y=crfj]{\systemdata}; 
        \addplot+ table[x=data,y=crfjtr]{\systemdata}; 
        \addplot+ table[x=data,y=neural]{\systemdata};
        \addplot+ table[x=data,y=gnb]{\systemdata};
        \legend{\crfm, \crfm+\clp{}, \neural, \gnb}
    \end{axis}
\end{tikzpicture}

    \caption{Accuracy on the (remaining) test set as we add annotated narratives to the training set. All methods benefit from the active learning approach, with \gnb{} displaying better performance due to its use of monolingual data in the first iterations, but the \crfm{} approach achieving the best results in the last iterations, eventually not even needing the cross-lingual type-level projections (+\clp{}).}
    \label{fig:system}
\end{figure*}
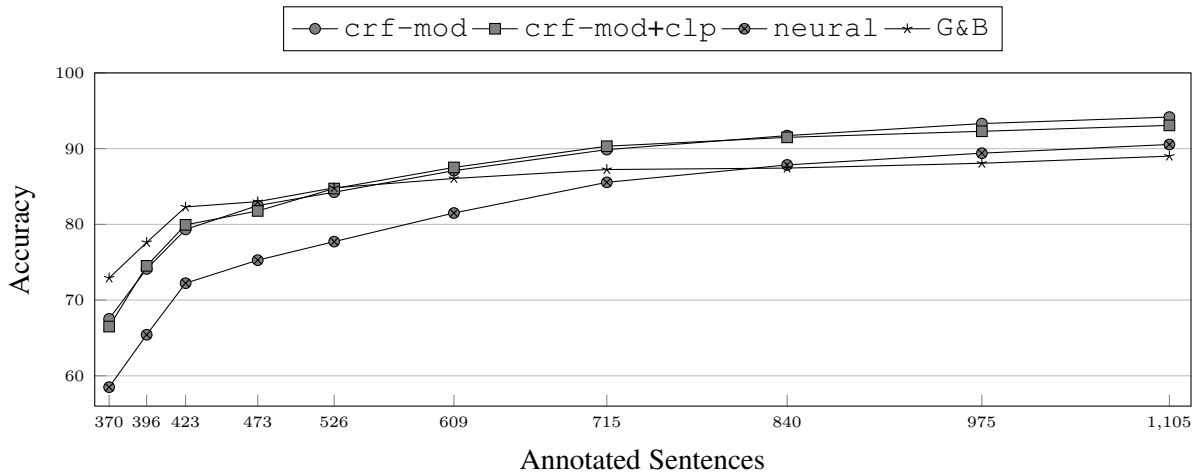

All methods display notable improvement as we added the annotated narratives to the training set. The performance trends are outlined in Figure~\ref{fig:system}. Firstly, it is notable that as the training set increases, the advantage of the model of \citet{garrette2013learning} that leverages monolingual data diminishes, compared to our simpler \crfm{} tagger, both with or without cross-lingual projected tags. Before the first iteration, the accuracy gap is $6.4$ percentage points in favor of \gnb{}. However, after adding around 4-5 narratives so that there are around~500 training sentences, our \crfm+\clp{} method surpasses the \gnb{} method and keeps improving. This is also outlined by the dashed line in Table~\ref{tab:narratives}. As we add more training instances, the accuracy of the \gnb{} method plateaus around~85\% and does not improve further.

Furthermore, after a couple more iterations, when more than~800 annotated sentences are available for training, the \crfm{} method without cross-lingual projected tags achieves higher accuracy than all others. We identify this point as the one where simple token-level supervision is efficient enough to outperform semi-supervised or transfer-learning approaches.

Finally, we observe that the accuracy of the \neural{} bi-LSTM approach that only uses the tagged corpus without further use of monolingual data, improves significantly as the training set increases. 
With only~370 training sentences, the gap between the \neural{} and the best method is more than~14 percentage points. With~1,100 training sentences, the accuracy gap diminishes to only~2 percentage points. 

\section{Cross-Validation}

Our end goal is to annotate the whole corpus with POS tags, as well as richer annotations. 
Towards that direction, our gold annotated test data could be used to train a higher quality POS tagger, which we will use to annotate the rest of the corpus. 
In Section~\S\ref{sec:active}, we found that including all but one annotated narratives for training, and testing on the last one (story-9) we were able to obtain an accuracy of more than~94.17\%.
In order to get a better estimation of how well a tagger trained on our gold data would work, we perform a cross-validation experiment, using \crfm{}, our best performing model.

For each cross-validation instance, one of the annotated narratives becomes the test set, and the rest will be included in the training set. This allows us to obtain an average performance over~10 instances. The average accuracy of the \crfm{} model is about~91.9\%, with a standard deviation of about~2 percentage points (minimum is~88.5\% on story 5, and maximum is~94.9\% on story 2).

The main obstacle to annotating the rest of the corpus with higher quality is out-of-vocabulary words. The combined vocabulary of the \uoi{} corpus and our~10 annotated narratives covers~16\% of the vocabulary of the~104 unannotated sentences (but~85\% of the total tokens). As part of our future work, we plan to incorporate word-level active learning in our annotation/correction scheme, similar to the approaches proposed by \citet{meng2017model}.

\section{Conclusion}

We presented a parallel corpus of 114 narratives on an endangered language, Griko, with translations in Italian. 
For now, a test set of 10 narratives is hand-annotated with Part-of-Speech tags, but in the future we will enrich the resource with annotations on the rest of the corpus, as well as with richer syntactic and morphological annotations. We also plan on contributing our corpus to the Universal Dependencies treebanks \cite{nivre2016universal} as Griko is absent from the supported languages.

We extensively evaluated several POS tagging approaches, and found that the method of \citet{garrette2013learning} can be combined with cross-lingual type-level projected tags, outperforming all other methods, with an accuracy of~72.9\%, when less than~500 sentences are available. 
As data was added in the training set in an active learning scenario, a simple feature-based CRF approach outperforms all other models, with accuracy improvements of over~21 percentage points and over~94\% accuracy on the last narrative. In fact, when more than~800 sentences are available for training, cross-lingual tag projections hurt performance.

The collected annotations from our test set could form the basis for training a high-accuracy POS tagger for Griko, so that we can expand the POS annotations to the rest of our corpus with only a small amount of noise. We aim to explore this direction in our future work, along with other active learning methods that require less human intervention. In addition, we plan to further enrich the annotations of our corpus with morphological tags similar to the \uoi{} corpus, that will provide even more insight in Griko and its usage.
When the full annotations of the corpus are completed, we plan to use statistical methods to study specific phenomena regarding the grammar and syntax of Griko.

Finally, and most importantly, we hope that the release of this corpus will spark further interest for computational approaches applied on endangered languages documentation and on under-represented languages in general. 

\paragraph{Acknowledgements} This work was generously partially supported by NSF Award 1464553. We are also grateful to the anonymous reviewers for their thougthful reviews and useful comments.

\bibliography{References}
\bibliographystyle{acl}

\end{document}